\documentclass[journal]{IEEEtran}
\usepackage{cite}
\usepackage{amsmath,amssymb,amsfonts}
\usepackage{graphicx}
\usepackage{textcomp}
\usepackage{physics}
\usepackage{xcolor}
\def\BibTeX{{\rm B\kern-.05em{\sc i\kern-.025em b}\kern-.08em
    T\kern-.1667em\lower.7ex\hbox{E}\kern-.125emX}}
\usepackage[utf8]{inputenc} 
\graphicspath{{../figures/}}
\usepackage{makecell}  
\usepackage[
  separate-uncertainty = true,
  multi-part-units = repeat
]{siunitx}
\usepackage{bm}        

\newcommand{\e}[1]{\ensuremath{\cdot 10^{#1}}}  
\newcommand*{\tran}{^{\mkern-1.5mu\mathsf{T}}}  

\newcommand\T{\rule{0pt}{2.6ex}}       
\newcommand\B{\rule[-1.2ex]{0pt}{0pt}} 
\newcommand{\figref}[1]{\figurename~\ref{#1}}
\newcommand{\tabref}[1]{Tab.~\ref{#1}}

\newcommand{\secref}[1]{Sec.~\ref{#1}}

\begin{document}

\title{Thermal Neural Networks: \\Lumped-Parameter Thermal Modeling With \\State-Space Machine Learning
\thanks{This work is supported by the German Research Foundation (DFG) under grant BO 2535/15-1. Computing time was provided by the Paderborn Center for Parallel Computing (PC\textsuperscript{2}).}
}

\author{
	\vskip 1em
	{
	Wilhelm Kirchgässner, 
	Oliver Wallscheid, \emph{Member}, \emph{IEEE}
	 and Joachim Böcker, \emph{Senior Member}, \emph{IEEE}
	}

	\thanks{
		
		{
		
		W. Kirchgässner and J. Böcker are with the Department of Power Electronics and Electrical Drives while O. Wallscheid is with the Automatic Control Department, Paderborn University, D-33095 Paderborn, Germany\\ (e-mail: kirchgaessner, wallscheid, boecker@lea.upb.de).
		}
	}
}

\maketitle

\begin{abstract}
With electric power systems becoming more compact and increasingly powerful, the relevance of thermal stress especially during overload operation is expected to increase ceaselessly.
Whenever critical temperatures cannot be measured economically on a sensor base, a thermal model lends itself to estimate those unknown quantities.
Thermal models for electric power systems are usually required to be both, real-time capable and of high estimation accuracy.
Moreover, ease of implementation and time to production play an increasingly important role.
In this work, the thermal neural network (TNN) is introduced, which unifies both, consolidated knowledge in the form of heat-transfer-based lumped-parameter models, and data-driven nonlinear function approximation with supervised machine learning.
A quasi-linear parameter-varying system is identified solely from empirical data, where relationships between scheduling variables and system matrices are inferred statistically and automatically.
At the same time, a TNN has physically interpretable states through its state-space representation, is end-to-end trainable -- similar to deep learning models -- with automatic differentiation, and requires no material, geometry, nor expert knowledge for its design.
Experiments on an electric motor data set show that a TNN achieves higher temperature estimation accuracies than previous white-/grey- or black-box models with a mean squared error of 3.18~K² and a worst-case error of 5.84~K at 64 model parameters.
\end{abstract}


\section{Introduction}
\label{sec:intro}
Lumped-parameter thermal networks (LPTNs) are a system of ordinary differential equations (ODEs) and a popular choice for various thermal modeling applications throughout different industry branches.
Most notably, their relatively low amount of model parameters lends themselves for real-time temperature estimation tasks in electric power systems, such as multichip power modules \cite{IaLu2014, BaMaGhi2016}; automotive-grade electric machines \cite{WaBo2016, BraHe2012, BoCaCo2015}, or even in civil engineering \cite{RaGoEa2013}. 
An accurate real-time temperature estimator is often highly desirable whenever a system's control performance depends on thermal state information of such parts that are not economically measurable on a sensor base, e.g., the permanent magnets in a synchronous electric machine or semiconductors in power electronic converters.
Their need is particularly urgent when overload operation can be set, which potentially harms the system's health through thermal stress.
Even with system temperatures being tracked by sensors, a redundant thermal model becomes more and more relevant against the background of increasing functional safety requirements.

Being a parameter-light model often makes the LPTN approach the only feasible choice, although established, computational heavy, thermal analysis tools like computational fluid dynamics (CFD) and finite-element modeling (FEM) based on partial differential equations (PDEs) usually give a more thorough and accurate picture of heat flows in any intricate system \cite{BoCa2009}.
Apart from thermal models, there are also real-time methods in the electric machine domain based on either invasive, high-frequency signal injection \cite{ReiFeYo2015, ReiFeMa2019} or precice observers that indirectly assess an electric machine's thermal condition from temperature-dependent electric parameters \cite{SpeWaBo2014}.
Both methods expose high system parameter sensitivity as well as operation-point-dependent accuracies, that might lead to undesired temperature estimation errors \cite{WaHuPe16}.
What is more, a considerable amount of domain knowledge is required to be incorporated in each system model.

\subsection{State-of-the-art Thermal Modeling}
\begin{figure}[tb]
    \centering
    \includegraphics[width=0.49\textwidth]{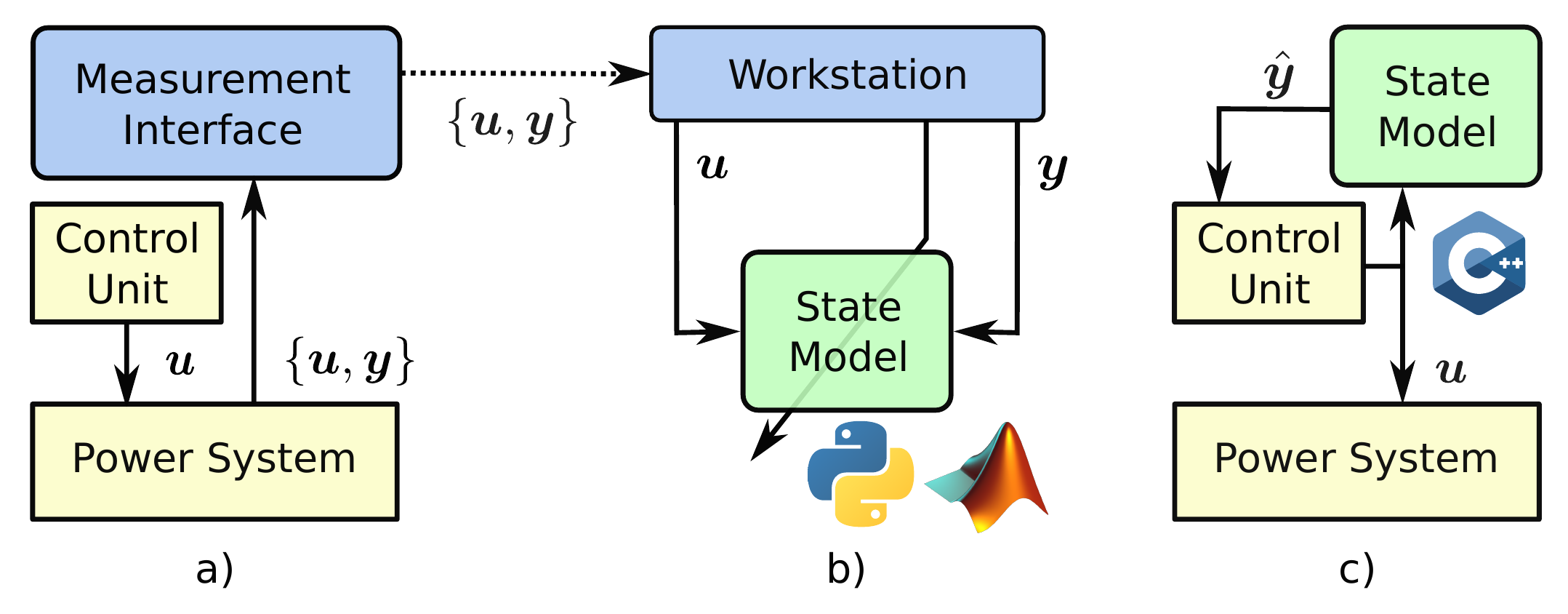}
    \caption{The data-driven state modeling process is depicted. a) The power system is either upgraded with measurement equipment or operated at a test bench, and reveals information about the system's input $\bm u$ and measured states $\bm y$ (e.g, interior temperatures). b) The recorded data set is used to tune a state model on estimating $\bm y$. c) The tuned model is deployed on an embedded system and informs the control unit.}
    \label{fig:scheme}
\end{figure}

Likewise, typical for an LPTN design of sufficient accuracy is the incorporation of expert knowledge from the application's domain, as well as geometry and material information from manufacturers.
These help to identify the parameter variation of thermal resistances and capacitances, yet the geometry is often approximated with simple shapes (cylinders, spheres) to further reduce the modeling complexity of also intricate component shapes.
What is more, power losses are usually even more difficult to grasp, and significant research effort is put into their modeling \cite{GeWaBo2020, QiScheDo2014}.
In order to overcome modeling errors imposed by simplified assumptions over the system, empirical measurement data (for instance, as collected on a test bench) is increasingly utilized and employed, e.g., in the form of look-up tables or for fine-tuning analytically found initial parameter values \cite{WaBo2016, WoeBraDre2020, GeWaBo2020}.
The idea is depicted in \figref{fig:scheme}.

Advancing the trend of fitting models on empirical data, machine-learning-based thermal modeling was proposed recently, where artificial neural networks (ANNs) are tuned to estimate temperatures of interest from a parallel stream of readily available sensor data, effectively harnessing statistical correlations between the latent temperatures and measurable quantities \cite{KiWaBo2020, KiWaBo2021_benchmark, LeeHa2020, ZhaGuOg2018}. 
These methods are completely detached from thermodynamic theory and rely solely on empirical data.
They neither necessitate expert knowledge nor geometry or material information for their design at the same level of accuracy as that of data-driven LPTNs.
However, an increased computational demand in real time is required due to additional feature engineering and a substantially increased amount of model parameters \cite{KiWaBo2021_benchmark}.
Moreover, learnt parameters of an ANN are not as interpretable as of an LPTN.
This implies that other systems, albeit similar to the one observed by the black-box model, would always require, again, empirical data for an appropriate temperature estimation unit.
Furthermore, initial conditions can not be set in black-box models as their internal states are abstract, if there are any at all.

\subsection{Lumped-Parameter Thermal Networks}
\begin{figure}[htb]
    \centering
    \includegraphics[width=0.19\textwidth]{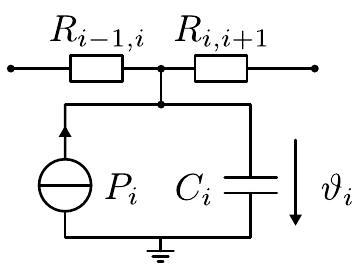}
    \caption{A typical thermal element consisting of a thermal capacitance, a power loss source, and multiple thermal resistances leading to other thermal elements. $\vartheta_i$ denotes the $i$-th element's average temperature.}
    \label{fig:thermal_element}
\end{figure}

Based on heat transfer theory, LPTNs are designed to estimate specific spatially separated system components' temperatures by representing heat flows through an equivalent circuit diagram, where parameters stand for thermal characteristics.
An appropriate LPTN structure is usually derived from the heat diffusion equation \cite{Bergman2007}:
\begin{equation}
	\rho c_\text{p} \pdv{\vartheta}{t} = p + \nabla\cdot(\lambda \nabla\vartheta),
	\label{eq:heat}
\end{equation}
where $\rho$ denotes the mass density, $c_\text{p}$ stands for the volume's specific heat at constant pressure, $\vartheta$ is the scalar temperature field, $p$ being the thermal energy generation from other energy sources (such as electric or chemical energy), $\nabla$ being the del operator across spatial dimensions, and $\lambda$ representing the potentially nonconstant thermal conductivity.
For steady-state heat transfer, a constant thermal resistance can then be derived analytically from geometry as well as boundary and initial conditions to describe one-dimensional heat flow that depends on the absolute temperature difference, much like current flow depends on potential differences in an equivalent electric circuit.
In the multi-dimensional case, converting \eqref{eq:heat} into a spatial finite-difference equation is often helpful.
\figref{fig:thermal_element} shows a thermal element, which is an LPTN's basic building block and represents one particular subdomain in the system.
This simplified diagram discretizes a system's state space and converts the governing PDEs into ODEs.

However, the parameters in \figref{fig:thermal_element} are commonly operation-point-dependent rather than being constant for most applications.
Moreover, especially convective and radiative heat transfer processes inherit highly nonlinear thermal characteristics making the transient definition of thermal parameters a challenging endeavor \cite{BoCaPa06, HiChiHo12}.
Usually, thermal parameters are thus denoted in dependence on a scheduling vector $\bm \zeta(t)$.
This vector contains not only measurement quantities that are available during operation but also all thermal states themselves, which will be discussed later.
%

In order to avoid incorporation of geometric specifications upfront, a general-purpose LPTN structure of the following form is considered as starting point for subsequent designs:
The $i$-th thermal element in a fully interconnected LPTN of $m$-th order describes the ODE
\begin{align}
\begin{split}
	C_i\big(\bm\zeta(t)\big) \dv{\vartheta_i}{t} &= P_i\big(\bm\zeta(t)\big) + \\
	& \sum_{j\in \mathcal{M} \setminus i}{\frac{\vartheta_j - \vartheta_i}{R_{i,j}\big(\bm\zeta(t)\big)}} + \sum_{j=1}^n{\frac{\tilde{\vartheta}_j - \vartheta_i}{R_{i,j}\big(\bm\zeta(t)\big)}},
\end{split}
\label{eq:lptn_ode}
\end{align}
where $C$ denotes the thermal capacitance, $P$ the power loss, $R_{i,j}$ the bidirectional thermal resistance between the temperatures at node $i$ and $j$, and $\vartheta$ the actual temperature at time $t$ with $\mathcal{M} = \{1,2, \dots, m\}$.
Moreover, temperatures that are measurable during operation (typically ambient and coolant temperatures) can be incorporated as further $n$ ancillary nodes merely consisting of an equivalent thermal source $\tilde{\vartheta}$.

It will prove beneficial for the following sections to rearrange \eqref{eq:lptn_ode} into a quasi-linear parameter-varying (quasi-LPV) state-space representation encompassing all nodes in matrix form:
\begin{align}
\begin{split}
	\dv{\bm x}{t} &= \bm A\big(\bm\zeta(t)\big)\bm x + \bm B\big(\bm\zeta(t)\big)\bm u\big(\bm\zeta(t)\big), \\
	\bm y &= \bm C\bm x,
\end{split}
	\label{eq:lptn_state_space}
\end{align}
with
\begin{align*}
	\bm x &= \bm\vartheta = [\vartheta_1 \dots \vartheta_{m}]\tran,\\
	\bm u &= [P_1\big(\bm\zeta(t)\big) \dots P_{m}\big(\bm\zeta(t)\big)\quad \tilde{\vartheta}_{1} \dots \tilde{\vartheta}_n]\tran,\\
	\bm A &\in \mathbb{R}^{m\times m},\quad \bm B \in \mathbb{R}^{m\times (m+n)},\quad \bm C \in \mathbb{R}^{q\times m},
\end{align*}
where $\bm A$, $\bm B$, and $\bm C$ are the system, input, and output matrix, respectively.
The output matrix $\bm C$ is typically a masking operation allowing the system output $\bm y$ to encompass fewer temperatures than are being modeled with $q \leq m$.
Identification of $\bm A\big(\bm\zeta(t)\big)$ and $\bm B\big(\bm\zeta(t)\big)$ is central to thermal modeling with LPTN structures.

Furthermore, as mentioned in the previous section, determining the dependence of $P_i$ on $\bm\zeta(t)$ is usually a hard task:
there are often only few sensor information (e.g., electrical current at the terminals of a power electronic converter or electric machine) from which a complex nonlinear spatial power loss distribution is to be inferred.
Finding this mapping is notoriously challenging in addition to identifying $\bm A\big(\bm\zeta(t)\big)$ and $\bm B\big(\bm\zeta(t)\big)$.

\subsection{State-Space Machine Learning}
Recently, a paradigm shift emerged promoting the combination of general-purpose machine learning (ML), stemming from computer science, and prior scientific knowledge in the form of PDEs commonly appearing in nature, instead of contemplating these as distinct problem sets.
Leveraging the potential of automatic differentiation \cite{GunesBaydin2018}, various data-driven approaches were presented for both, discovery and solution of nonlinear PDEs \cite{Rudy2017, CheRuBeDu2018, RaPeKa2019, Rackauckas2020}: 
a prominent idea is the physics-informed ANN (PINN), which adds any PDE of interest to a standard ANN's cost function \cite{RaPeKa2019}.
While a fitted PINN has been shown to respect PDEs with certain boundary and initial conditions, their convergence properties are not fully understood yet, and, more notably, their architecture is unaltered compared to usual ANNs.
This makes a PINN as noninterpretable as typical black-box models.

Conversely, neural ordinary differential equations denote initial value problems of the form
\begin{align}
	\dv{\bm{x}}{t} = \bm g_{\bm\theta}(\bm x, \bm u, t),
	\label{eq:node}
\end{align}
with $\bm x$ denoting a system's state at time $t$, $\bm u$ the system's input, and $\bm g_{\bm\theta}$ being a neural network function parameterized with $\bm\theta$ \cite{CheRuBeDu2018}.
Although this is a first step to applying an outer structure to an ANN that imitates the nonlinear workings of ODEs, it relies too heavily on the universal approximation theorem \cite{HoSti1989}, and prior scientific knowledge of fundamental mechanisms are not directly incorporated.

This shortcoming is addressed by universal differential equations (UDEs), which extend the ODE case \eqref{eq:node} to
\begin{align}
	\dv{\bm{x}}{t} = \bm f(\bm x, \bm u, t, \bm g_{\bm\theta}(\bm x, \bm u, t)),
	\label{eq:uode}
\end{align}
where $\bm f$ represents arbitrary known mechanisms of the underlying physical problem in which difficult to model parts are approximated by any universal approximator $\bm g_{\bm\theta}$ \cite{Rackauckas2020}.
Note, however, that except for some concrete examples with very specific ANN constellations there is no rule-based procedure proposed on how to receive an appropriate UDE given a nonlinear system.
Due to the broadly outlined definition of an UDE, most modeling problems still remain to be solved by experts in the corresponding domain.

\subsection{State-Space Neural Networks}
The neural ODE \eqref{eq:node} strongly resembles the state-space neural network (SSNN), which was proposed within the control theory domain over two decades ago \cite{Rivals1996}.
SSNNs were an early attempt to combine ANNs with state-space representations of general nonlinear, dynamic systems, in order to leverage well established stability and convergence theorems with the nonlinear modeling capability of ANNs.
They describe the difference equation
\begin{align}
\begin{split}
	\hat {\bm x}[k+1] &= \bm g_{\bm\theta}(\hat{\bm x}, \bm u, k),\\
	\hat {\bm y}[k] &= \bm h_{\bm\theta}(\hat{\bm x}, \bm u, k),
	\label{eq:ssnn}
\end{split}
\end{align}
where $\bm g_{\bm\theta}$ and $\bm h_{\bm\theta}$ are nonlinear functions represented by potentially the same ANN with parameters $\bm \theta$ at discrete time $k$.
Obviously, the mere difference to neural ODEs \eqref{eq:node} is the differential equation form, but, again, the universal approximation theorem is relied on completely here.
Following this, most contributions deriving from SSNNs employ some form of recurrent ANNs (RNNs)
\begin{equation}
	\bm g_{\bm \theta}(\hat{\bm x}, \bm u, k) = \bm\sigma(\bm W_\text{r}\hat{\bm x}[k] + \bm W_\text{h}\bm u[k] + \bm b),
	\label{eq:rnn}
\end{equation}
with $\bm W_{\{\text{r,h}\}}$ being the recurrent and forward weight matrices, $\bm b$ being the bias vector, and $\bm \sigma$ denoting the RNN's activation function \cite{ssNN98, AbbWe2008}.
While the computer-science-based movement on ML estimation of state derivatives does not restrict the topology of the state estimator to any form (e.g., certain number of layers, standard vs. convolutional ANNs, etc.), SSNN derivatives are often proposed with those settings fixed.
This might stem from the desire to be compatible to existing control theorems, yet it could also originate from the lack of automatic differentiation tools, which render training of large ANNs possible and emerged just a decade ago during the revived interest into neural networks.

However, the idea of estimating $\hat {\bm x}[k+1]$ under a certain time discretization is central to this article as well, albeit here the chosen topology is more sophisticated than \eqref{eq:rnn} and tailored to heat transfer processes.
As will be pointed out in the next section, this contribution is motivated by \eqref{eq:uode} and informed by \eqref{eq:lptn_state_space}.

\subsection{Contribution}
In this work, the thermal neural network (TNN) is proposed, which facilitates physically reasoned heat transfer mechanics and can be applied to any system that exposes component temperatures of particular interest.
Unifying the concepts of LPTNs and UDEs, a TNN is universally applicable across all relevant engineering domains that deal with heat transfer.
In particular, it integrates the following advantages into one modeling approach:
\begin{itemize}
	\item Physically interpretable model states are featured, which make initial condition adjustment simple.
	\item TNNs are end-to-end differentiable, which enables fast, parallelized, and distributed training scenarios with automatic differentiation frameworks.
	\item Neither expert knowledge, nor material or geometry information are required for the design.
	\item Being completely data-driven, parasitic real-world effects are considered through observational data.
	\item A quasi-LPV system is identified where unknown relationships between scheduling variables and system matrices are automatically discovered.
	\item Even relatively small model sizes achieve high estimation accuracies, which facilitates real-time capability.

\end{itemize}
This article begins with the working principle of a TNN, followed by the demonstration of its modeling capacity through cross-validation on a representative data set.
A kernel-based hyperparameter optimization is described, and chances for model reduction are shown.
Finally, the TNN's initial condition adjustment and its recovering properties are demonstrated.
To assist related work all code is published on GitHub\footnote{https://github.com/wkirgsn/thermal-nn}.

\section{Thermal Neural Networks}
Circumventing the intricate and error-prone design effort revolving around identification of the functions that are subject to the thermal equivalent parameters, the following architecture is leveraging the high nonlinear modeling capacity of ANNs.
It learns the existing physical relationship between temperatures and other sensor information without geometry or material information - just from measurement data and the general structure of an LPTN.
Building on the concept of \eqref{eq:uode}, the thermal neural network (TNN) reformulates \eqref{eq:lptn_ode} after a first-order Euler discretization to 

\begin{align}
\begin{split}
	 & \hat\vartheta_{i}[k+1] = \hat\vartheta_{i}[k] + T_\text{s} \kappa_i[k] \bigg(\pi_i[k] + \\	 
	 & \sum_{j \in \mathcal{M} \setminus i}{(\hat\vartheta_{j}[k] - \hat\vartheta_{i}[k]) \gamma_{i,j}[k]} +
	  \sum_{j=1}^n{(\tilde{\vartheta}_j[k] - \hat{\vartheta}_i[k]) \gamma_{i,j}[k]} \bigg),
\end{split}
\label{eq:tnn}
\end{align}

with $\hat\vartheta_{i}[k]$ denoting the $i$-th node's normalized temperature estimate at discrete time $k$, $T_\text{s}$ being the sample time, and $\kappa_i$, $\pi_i$ as well as $\gamma_{i,j}$ denoting arbitrary feed-forward ANN outputs dependent on $\bm\zeta[k] = [\tilde{\bm{\vartheta}}\tran \quad \hat{\bm{\vartheta}}\tran \quad \bm\xi\tran]\tran$, which in turn consists of the ancillary temperatures $\tilde{\bm{\vartheta}}[k]$, the temperature estimates $\hat{\bm{\vartheta}}[k]$, and additional observables $\bm\xi[k] \in \mathbb{R}^o$.
More specifically, $\bm\gamma(\bm\zeta)$ approximates all thermal conductances between components of interest, $\bm\pi(\bm\zeta)$ all power losses within these components, and $\bm\kappa(\bm\zeta)$ determines all inverse thermal capacitances.
Common observables are, e.g., the current and voltage vector as well as the motor speed in an electric drive train.

\begin{figure}[htb]
    \centering
    \includegraphics[width=0.49\textwidth]{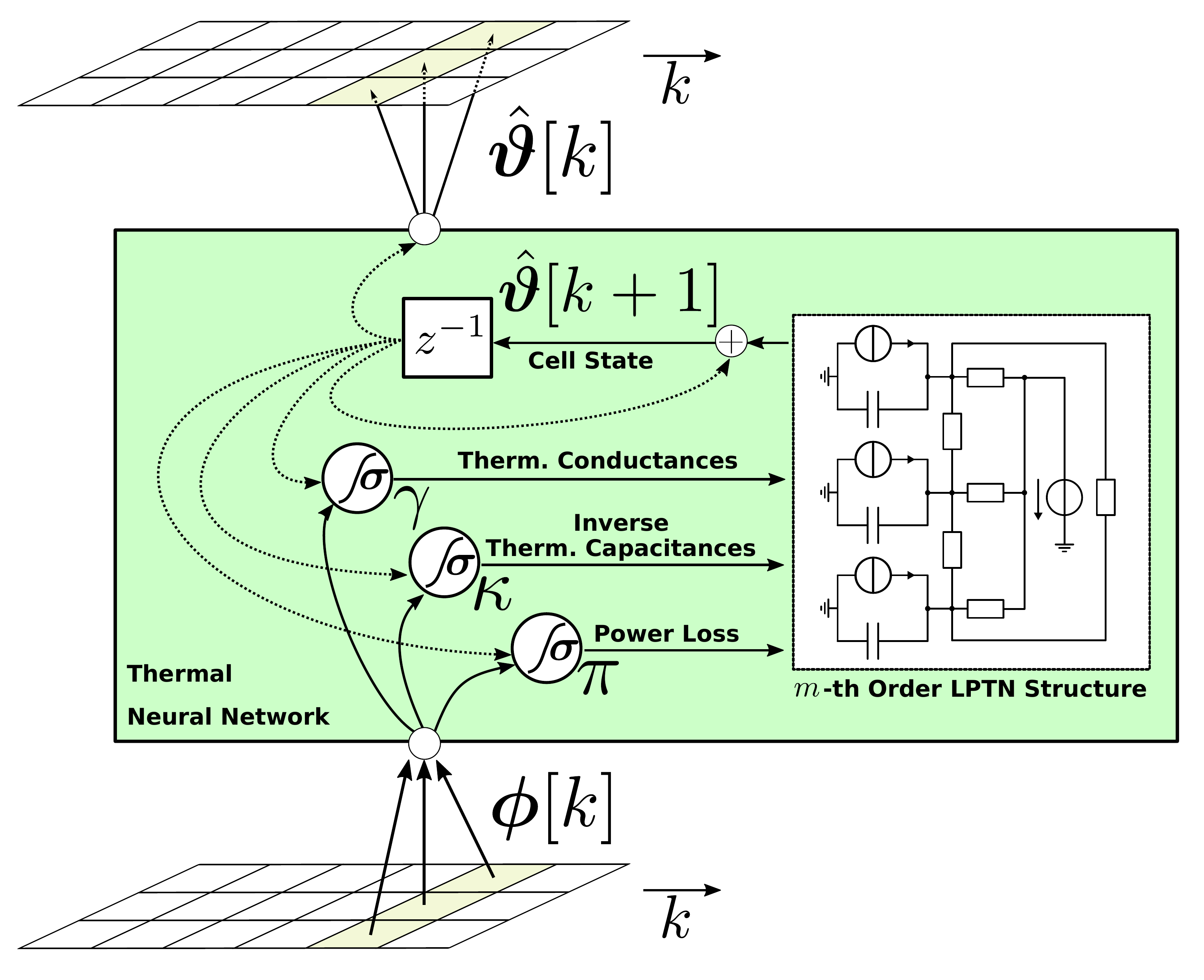}
    \caption{A TNN considers all input features of the current time step $\bm \phi[k]$ to estimate the next time step's temperatures $\bm \vartheta[k+1]$. This estimate acts as cell information within the TNN to further inform the three function approximators $\bm\gamma$, $\bm\kappa$, and $\bm\pi$. An LPTN structure of third order with an ancillary thermal source is exemplarily depicted.}
    \label{fig:tnn_concept}
\end{figure}

Without loss of generality, $\bm\kappa(\bm\zeta)$ can usually be reduced to just end-to-end trainable constants $\bm\theta_\text{c}$ in the form 
\begin{align*}
\bm\kappa = 10^{\bm{\theta_\text{c}}},\quad \bm\theta_\text{c} \in \mathbb{R}^{m},
\end{align*}
whereas 
\begin{align*}
\bm\pi:& \mathbb{R}^{m+n+o} \to \mathbb{R}^{m}, \\
\bm\gamma:& \mathbb{R}^{m+n+o} \to \mathbb{R}^{(m+n)(m+n-1)/2}
\end{align*}
are distinct ANNs.
This reduction corresponds to a simplification through the lumped capacitance method, which is feasible in many engineering applications \cite{Bergman2007}.
The TNN concept is sketched in \figref{fig:tnn_concept}.

Hence, except for $\bm\kappa$, the general algorithmic shape is described by the multilayer perceptron (MLP) algorithm fed by $\hat{\bm{\vartheta}}$ and $\bm\phi = [\tilde{\bm \vartheta}\tran \quad \bm\xi\tran]\tran$:
\begin{align}
\begin{split}
	\bm h^{(0)}[k] &= \bm\sigma^{(0)}(\bm W_\text{r}\hat{\bm \vartheta}[k] + \bm W_\text{h}^{(0)}\bm\phi[k] + \bm b^{(0)}), \\
	\bm h^{(l)}[k] &= \bm\sigma^{(l)}(\bm W_\text{h}^{(l)}\bm h^{(l-1)}[k] + \bm b^{(l)}),\quad \forall \, l > 0, \\
	\bm g_{\bm \theta}[k] &= \bm h^{(L-1)}[k],
\end{split}
\label{eq:mlp}
\end{align}
with $\bm\sigma^{(l)}(\cdot)$ denoting the nonlinear activation function at layer $l$, and $\bm\theta = \{\bm W_\text{r}, \bm W_\text{h}^{(l)}, \bm b^{(l)} : l \in [0,L-1]\}$ describing the trainable parameters that exist independently for both, $\bm\pi$ and $\bm\gamma$.
Note that if one finds the assumptions under the lumped capacitance method to be too simplistic for the task, lifting $\bm\kappa$ to the MLP form would be an appropriate first measure.
Rearranging \eqref{eq:tnn} into matrix form similar to \eqref{eq:lptn_state_space} and substituting $(\bm \kappa, \bm\pi, \bm\gamma)$ with \eqref{eq:mlp} will reveal the relationship between the scheduling variables and the quasi-LPV system matrices as modeled by a TNN.

In ML terminology, the TNN is an RNN with one single layer, whose cell consists of three (arbitrary complex) ANNs.
It has a single cell state vector being the estimated temperatures and the TNN's output for the next time step, much like an SSNN with $\bm C$ being the identity matrix.
Dedicating three different function approximators to the thermal characteristics of an LPTN and acting on the discretized ODE rather than on the difference equation is what distinguishes the TNN from an SSNN.

\subsubsection{Training Remarks}
It is well known that training RNNs with the error backpropagation method suffers from the vanishing and exploding gradient problem \cite{PaMi2012}.
The problem defines the phenomena of gradients soon becoming very large or very close to zero the longer the sequence on which gradients are accumulated for the subsequent weight update.
This is especially severe for arbitrary long sequences on which thermal models conduct estimates.
Early on in ML literature, this problem was curbed with the introduction of memory blocks, like the long short-term memory (LSTM) block, that would expand each RNN neuron \cite{GeSch1999}.
However, the general-purpose topology of an LSTM is in conflict with the physically motivated TNN, such that a TNN has to overcome this hurdle with other methods, e.g., gradient normalization, clipping, and truncated backpropagation through time (TBPTT) \cite{WiPe90}.

\subsubsection{Physical Interpretation}
Since no information about the geometry or material of the system is assumed to be known a priori, a fully-connected LPTN denotes the starting point of the TNN design, i.e., $\bm A$ and $\bm B$ are dense matrices.
For a thermal model that is modeling $m+n$ different temperatures, this translates to $(m+n)(m+n-1)/2$ thermal conductances and, thus, a parameter size complexity of $\mathcal{O}((m+n)^2)$.
In practice, however, several components within a system are often sufficiently detached such that the heat transfer between them is neglegible.
Featuring physically interpretable states, a fitted TNN can evidence such prunable connections, that can be removed in a subsequent TNN design iteration.
This parameter reduction method is presented in \secref{sec:pruning}.

\subsubsection{Domain Knowledge}
If domain knowledge is available, this can be incorporated in several ways: for instance, by means of constraining the choice of topology defining parameters like the activation functions of the output layers of $\bm\gamma$, $\bm\kappa$, and $\bm\pi$, or by means of feature engineering or elimination per thermal parameter estimator.
Without loss of generality, it can be recommended to apply the $\ell_1$-norm on all the ANNs' outputs, as negative thermal conductances, capacitances, and power losses are rarely physically plausible.

\subsubsection{Initial Condition Adjustment}
As is usual for state-space models, the TNN structure asks for an initial temperature estimate.
In simulations and training, these initial values can be set with the ground truth values, while in a field application the ambient temperature is a good approximation.
Note that this initial conditioning is not possible for black-box models which have completely abstract parameter structures.
The ability to set the initial condition is thus to be seen as a major advantage in favor of models representing a dynamic state space.
The severeness of a biased initial guess is investigated in \secref{sec:detuned_init}.


\section{Experiments}
In order to demonstrate the TNN concept's efficacy, several experiments are conducted on a publicly available dataset\footnote{https://www.kaggle.com/wkirgsn/electric-motor-temperature}, which was also already utilized in \cite{KiWaBo2019_2, GeWaBo2020, KiWaBo2020, KiWaBo2021_benchmark}.
This data set represents $\SI{185}{\hour}$ of multivariate measurements sampled at $\SI{2}{\hertz}$ from a \SI{52}{\kilo\watt} three-phase automotive traction permanent magnet synchronous motor (PMSM) mounted on a test bench, which is torque-controlled while its motor speed is determined by a speed-controlled load motor ($\SI{210}{\kilo\watt}$) and fed by a 2-level IGBT inverter.
All measurements were recorded by dSPACE analog-digital-converters (ADCs) which have been synchronized with the control task.
\tabref{tab:motor_param} compiles the most important test bench parameters.  

\begin{figure}[htb]
    \centering
    \includegraphics[width=0.49\textwidth]{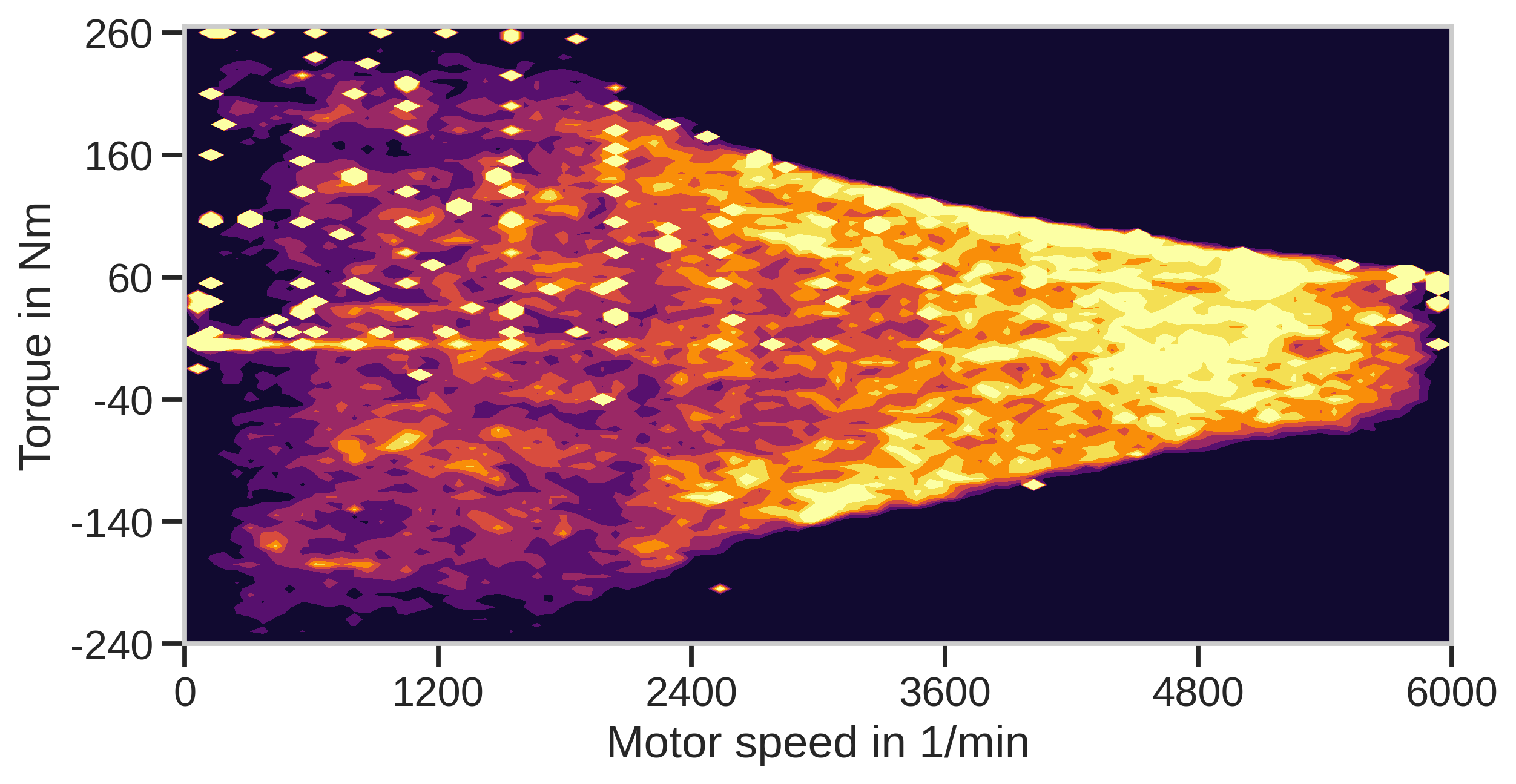}
    \caption{An interpolated 2D-histogram of visited operation points is shown in the motor-speed-torque-plane. Brighter areas were visited more often.}
    \label{fig:2d_hist}
\end{figure}

The contour plot in \figref{fig:2d_hist} depicts the excitation within the data set, which is characterized by constant and randomized excitation likewise.

\begin{table}[ht]
	\caption{Test bench parameters}
	\label{tab:motor_param}
	\centering
	\begin{tabular}{l|c|c}
	\hline\hline
	\textbf{Motors}				& device under test & load motor\T\\
	Type							 & PMSM  & induction machine\\
	Power rating (nom./max.)	    & $\SI{22/52}{\kilo\watt}$  & $\SI{160/210}{\kilo\watt}$\\
	Voltage rating			    & $\SI{177}{\volt}$ & $\SI{380}{\volt}$\\
	Current rating (nom./max.)	& $\SI{110/283}{\ampere}$	& $\SI{293/450}{\ampere}$\\
	Torque rating (nom./max.)	& $\SI{110/250}{\newton\meter}$ & $\SI{380/510}{\newton\meter}$ \\
	Pole pair number             & 8              		& 1 \\	
	Cooling						& water-glycol & forced air \\
	Thermocouples				& type K - class $1$ & - \B\\
	\hline
	\textbf{Inverter} & \multicolumn{2}{c}{3$\times$SKiiP 1242GB120-4DW} \T\\
	Typology                  & \multicolumn{2}{c}{voltage source inverter} \\
	        									& \multicolumn{2}{c}{2-level, IGBT} \\          
	Inverter interlocking time          &  \multicolumn{2}{c}{$\SI{3.3}{\micro\second}$} \B\\ \hline					
	\textbf{Controller hardware} & \multicolumn{2}{c}{dSPACE} \T\\
	Processor board              & \multicolumn{2}{l}{DS1006MC, 4 cores, 2.8\,GHz} \B\\ \hline\hline
	\end{tabular}
\end{table}	

\begin{figure}[tb]
    \centering
    \includegraphics[width=0.49\textwidth]{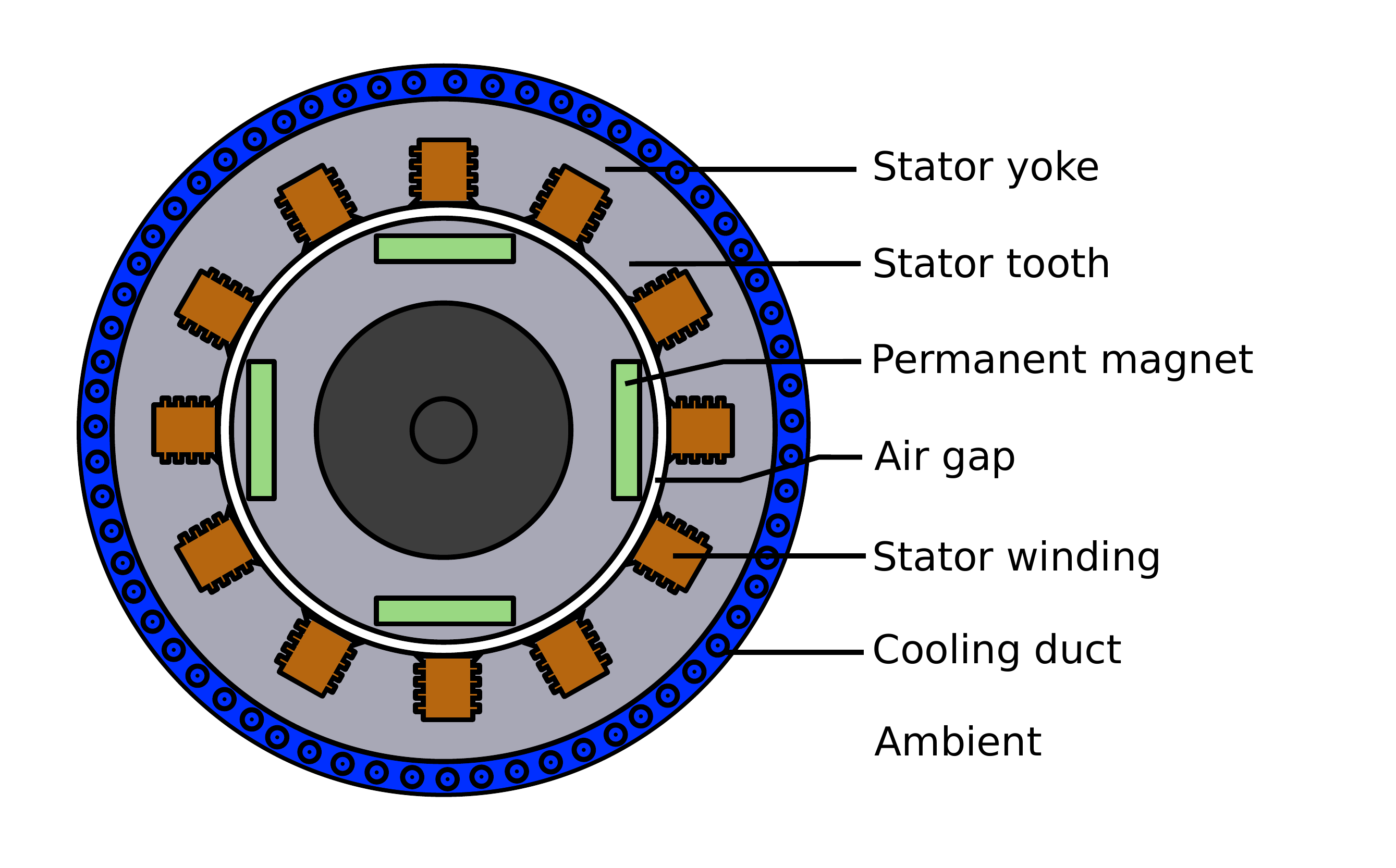}
    \caption{Exemplary and simplified PMSM cross-section with two pole pairs}
    \label{fig:motor_schnitt}
\end{figure}

The measured permanent magnet surface temperature as averaged across four sensors around a single permanent magnet is assumed to be representative for the rotor temperature, and is transmitted with a telemetry unit.
For all experiments, sensor information is classified in the following way:
\begin{align*}
	\bm \xi &= [u_\text{s}\quad i_\text{s}\quad \omega_\text{mech}]\tran, \\
	\tilde{\bm \vartheta} &= [\vartheta_\text{a}\quad \vartheta_\text{c}]\tran, \\
	\bm x &= \bm y = [\vartheta_\text{PM}\quad \vartheta_\text{SY}\quad \vartheta_\text{ST}\quad \vartheta_\text{SW}]\tran,
\end{align*}
with $u_\text{s}$ and $i_\text{s}$ being the voltage and current vector norm, respectively; $\omega_\text{mech}$ denoting the mechanical angular frequency; and $\vartheta_\text{\{a, c, PM, SY, ST, SW\}}$ representing temperatures at the ambient, coolant, permanent magnet, stator yoke, stator tooth, and stator winding, respectively.
An exemplary cross-section of a PMSM is sketched in \figref{fig:motor_schnitt} highlighting the sensors' geometric arrangement.
$\tilde{\bm \vartheta}$ is assumed to be the system's boundary, i.e., no heat transfer to other environmental temperatures are contemplated. 

As is usual for ANN training, all quantities are normalized: $i_\text{s}$ is divided by \SI{100}{\ampere}, $u_\text{s}$ by \SI{130}{\volt}, $\omega_\text{mech}$ by $2\pi\cdot\SI{6000}{\per\minute}$, and all temperatures by \SI{100}{\degreeCelsius}.
No further feature engineering is employed, which would be common for pure black-box alternatives where $\bm \phi$ would effectively grow to over $100$ elements \cite{KiWaBo2021_benchmark}.
The lack thereof can be seen as a computational advantage in favor of TNNs.

All models are evaluated for their estimation performance by the mean squared error (MSE) between estimated and ground truth trajectory, averaged across the targeted temperatures.
Proper cross-validation is ensured and described below.
Throughout this work, Tensorflow 2 \cite{tf2} is used to train and investigate TNNs.

\subsection{Automated Architecture Optimization}
Since a TNN only loosely defines the topological extent of its three thermal parameter estimators $\bm\gamma$, $\bm\kappa$, and $\bm\pi$, there is much room for hyperparameter optimization (HPO).

\subsubsection{Search Intervals}
Amongst topological factors, optimization-related hyperparameters are optimized as well.
An overview of all hyperparameters, their search intervals, and the found optima, is compiled in \tabref{tab:hpo}.
Generally, interval bounds were chosen to restrict parameter size growth into dimensions that would not be reasonably real-time computable anymore.

\begin{figure}[htb]
    \centering
    \includegraphics[width=0.49\textwidth]{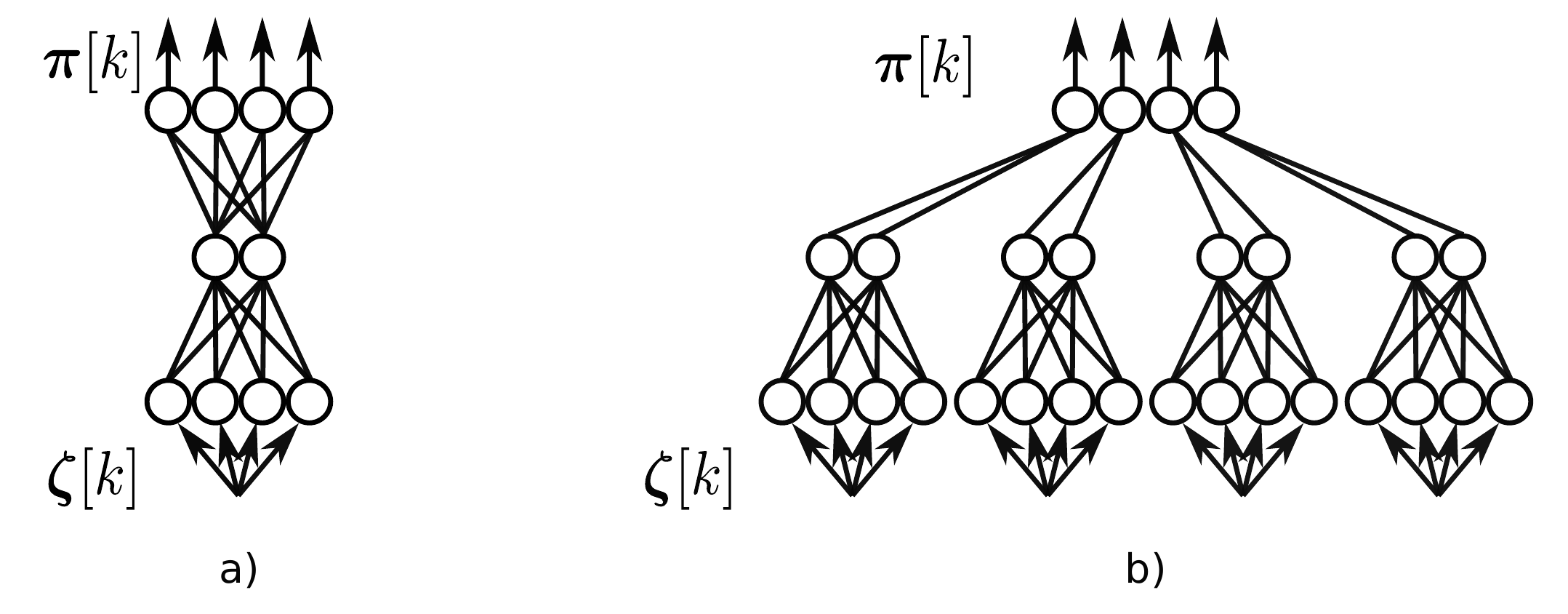}
    \caption{The HPO is given the opportunity to multiply the amount of model parameters in $\bm\pi$ (from (a) to (b)) by dedicating single MLP structures to each targeted component. This is a technique that has been proven useful in \cite{KiWaBo2020}. Given the higher amount of thermal conductances compared to targets, this scheme is not contemplated for $\bm\gamma$.}
    \label{fig:dedicated_branches}
\end{figure}

The set of activation functions $\alpha$ comprises those common in literature and the biased Elu $\sigma_\text{biasedElu}(\cdot) = \sigma_\text{elu}(\cdot) + 1$, in order to produce nonnegative output.
$P_{||}$ represents a flag that signals the expansion of $\bm\pi$ into dedicated MLP branches - one per target component (compare \figref{fig:dedicated_branches}).
The TBPTT length $\tilde{L}_\text{TBPTT}$ denotes how many forward passes are accumulated and averaged across for each backward pass during training with error backpropagation.
Splitting all measurement profiles into subsequences of a certain length $\tilde{L}_S$, where each subsequence is treated as an independent record, is indicated by the flag $S$.

\begin{table*}[tb]
        \renewcommand{\arraystretch}{1.8}
        \centering
        \caption{Hyperparameter search intervals and results}
	    \begin{tabular}{l l c c}
	        \hline\hline
	        
	        Symbol & Hyperparameter & Search interval & Optimum \T\B\\
	        \hline
	        $L_{\{\pi,\gamma\}}$ & No. hidden layers in $\{\bm\pi, \bm\gamma\}$ & [$1$, $3$]  & $\{3; 3\}$  \\ 
	        $n_{\{\pi,\gamma\}}^{(l)}$ & No. hidden units per layer in $\{\bm\pi, \bm\gamma\}$ & [$2$, $128$] &  $\{109\to3$; $2\to2\}$ \\
			$\sigma_{\{\pi,\gamma\}}^{(l)}$ & \makecell[l]{Activation function\\per layer in $\{\bm\pi, \bm\gamma\}$} &\makecell{Sigmoid, tanh, linear, ReLU,\\ biased Elu, sinus} & \makecell{$\{$ReLU$\to$sigmoid$\to$sigmoid;\\Tanh$\to$sinus$\to$biased Elu$\}$} \\
			$\alpha_{\{\pi,\gamma\}}^{(l)}$ &  $\ell_2$ regularization rate per layer in $\{\bm\pi, \bm\gamma\}$  & [$10^{-9}$, $10^{-1}$] & $\{10^{-8}\to10^{-8}$; $5.5\e{-6}\to$ $5.3\e{-8} \}$\\
			$P_{||}$ & Dedicated branches in $\pi$ & [True, False] & False \\  
	        $\eta$ & Initial learn rate & [$10^{-5}$, $1$]& $11\e{-3}$ \\
	        $\tilde{L}_\text{TBPTT}$ & TBPTT length in samples& [$8$, $2048$]  & $1227$ \\
	        $S$ & Split profiles into subsequences & [True, False] & False \\
	        $\tilde{L}_S$ & Subsequence length in hours & [$0.5$, $4$] & - \\
	        $\beta$ & Optimizer & Adam, Nadam, Adamax, SGD, RMSProp & Nadam \\
	        \hline\hline
	    \end{tabular}
	    \label{tab:hpo}
\end{table*}

\subsubsection{Cross-Validation Scheme}
\label{sec:cv}
The biggest stumbling block during the endeavor to achieve high estimation accuracies in data-driven modeling may be overfitting.
It explains the fact that, during training, any model of some expressivity can start to remember noise in the data in order to further minimize the underlying loss.
Appropriate cross-validation schemes need to be determined a priori in order to avoid such too optimistic generalization error assertions.

\begin{figure}[tb]
    \centering
    \includegraphics[width=0.49\textwidth]{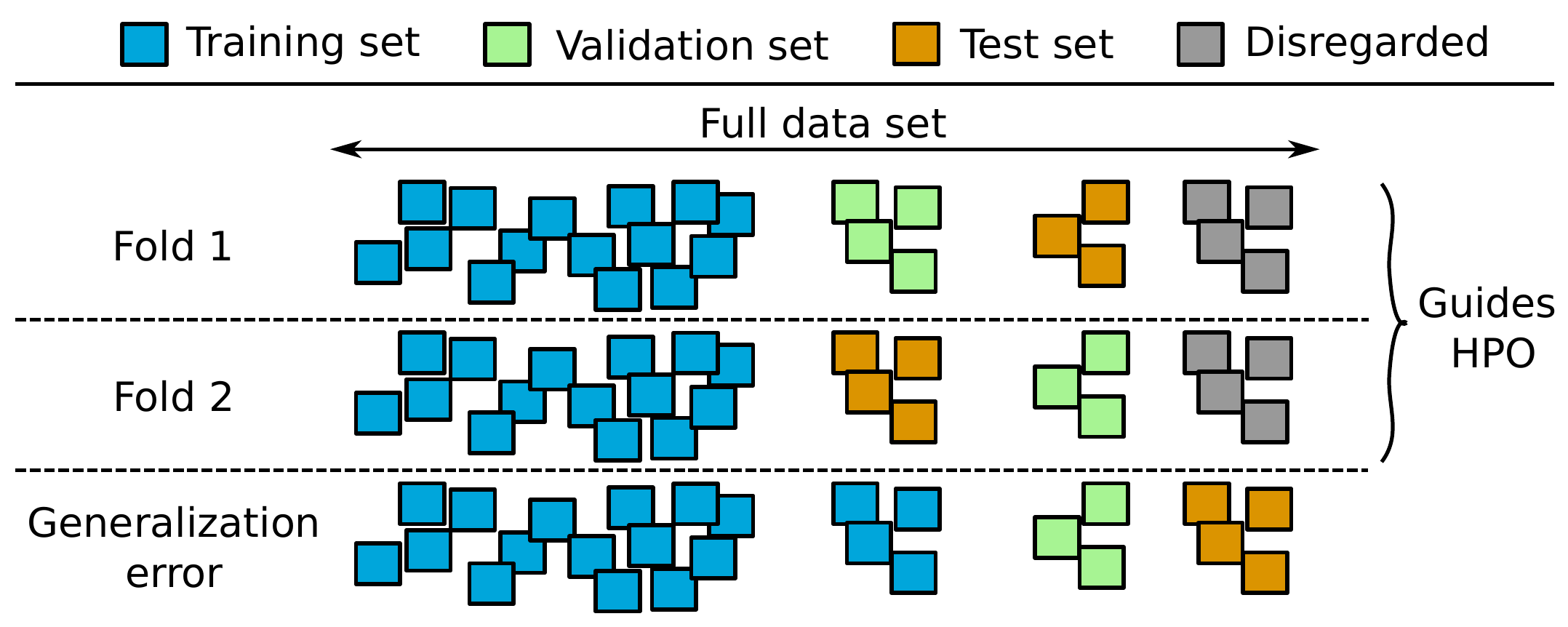}
    \caption{Cross-validation strategy where each square represents a measurement profile}
    \label{fig:cv}
\end{figure}

In this work, the investigated data set consists of several measurement records of different lengths.
A common approach is to split those records into different sets dedicated for either training or validating the model.
Following scheme is applied:
First, no record should be present in both training and testing sets at the same time, in order to preclude exploitation of record-specific anomalies.
Second, two so-called folds of mutually-exclusive sets are defined comprising a number of profiles that either act as validation or test set depending on the current iteration.
Both validation and test set are not observed by the model during training.
While the validation set is evaluated after each training epoch to apply early stopping, the test set is evaluated only once after training to receive an overall score.
The eventual score will be the average across both folds.
Last, considering the HPO being another source of overfitting but for hyperparameters, another set of profiles is excluded from training and validation during the HPO, and is called the generalization set, see \figref{fig:cv}.
Having found optimal hyperparameters, validating a model against this last set will give true generalization errors.

Furthermore, training outcome of ANNs is sensitive to their random initialization.
The same set of hyperparameters might end training at very different performance values depending on the first weight values being fortunate or not.
For this reason, during HPO, each training fold is repeated ten times with different random seeds and the score is then averaged across in order to assess the mean quality of a certain hyperparameter set.

\subsubsection{Search Results}
\begin{figure}[tb]
    \centering
    \includegraphics[width=0.49\textwidth]{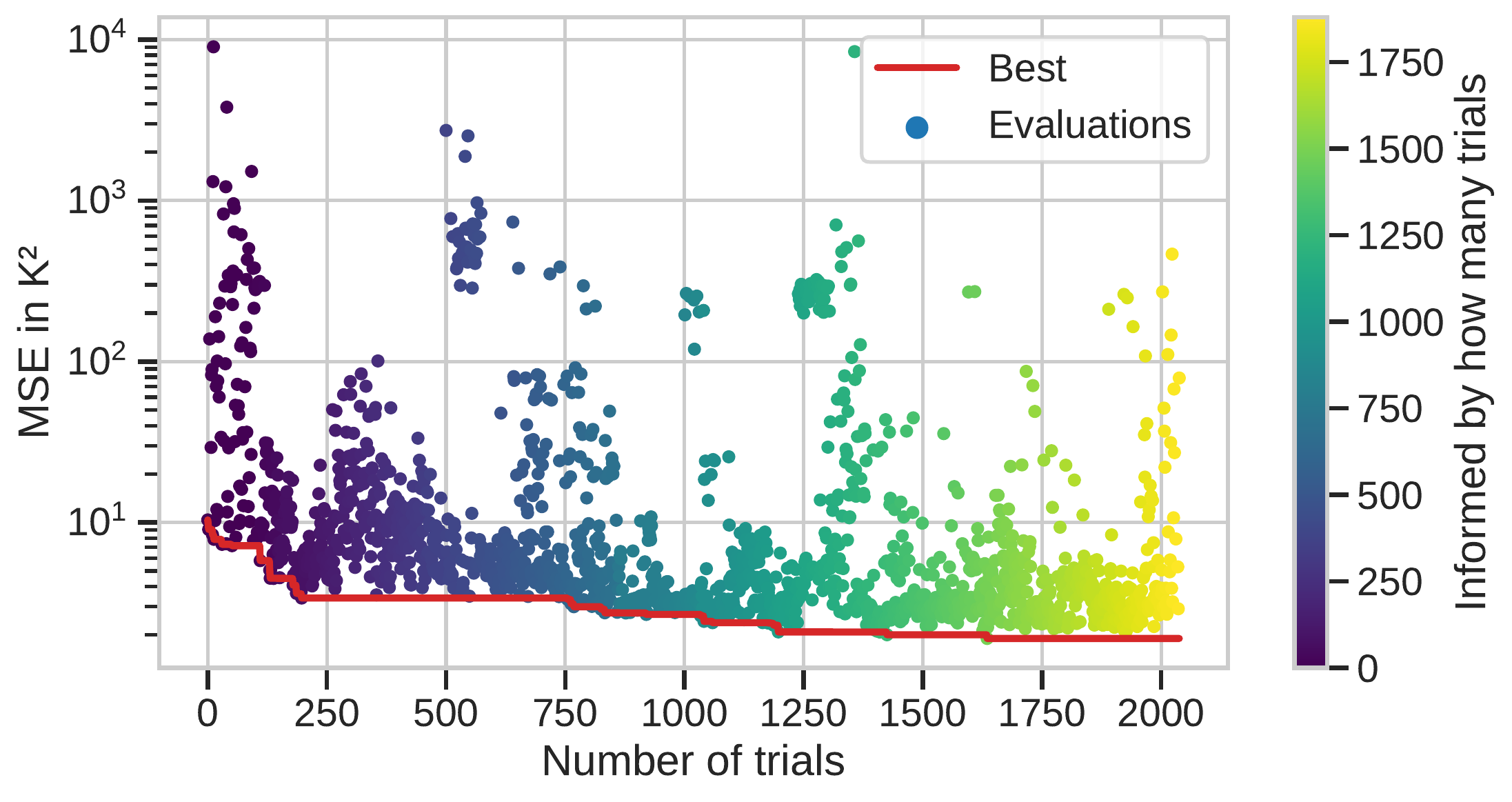}
    \caption{The HPO error trend over trials on the electric motor dataset is shown. Due to parallel execution, subsequent evaluations are not necessarily strictly-linearly informed by previous evaluations. Each trial denotes the average across ten random seeds on each of two test folds. The best MSE found is $\SI{1.9}{\kelvin\squared}$.}
    \label{fig:tnn_hpo_1}
\end{figure}

\begin{figure*}[tb]
    \centering
    \includegraphics[width=\textwidth]{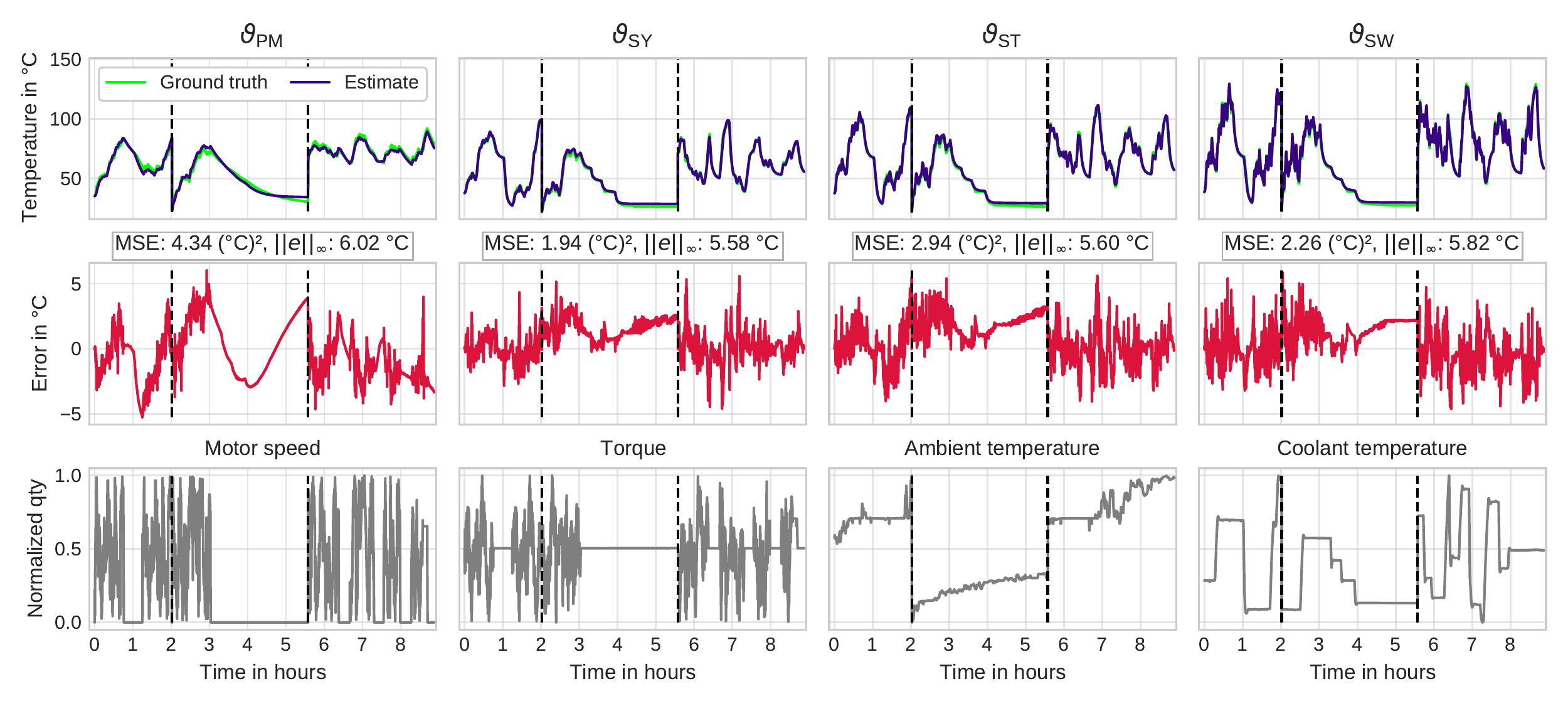}
    \caption{The performance of the found optimum on the generalization set is depicted. Dashed vertical lines separate independent profiles. Some selected sensor quantities are depicted at the bottom for reference.}
    \label{fig:best_genset_performance}
\end{figure*}

The HPO's sampling strategy is that of Optuna's tree-structured Parzen estimator (TPE) \cite{optuna_2019, BeBaBe2011}.
Hyperparameter sets were independently fetched and evaluated in parallel by $60$-$80$ workers on a high-performance computing cluster.
The HPO error trend is shown in \figref{fig:tnn_hpo_1}.
The overall best hyperparameter set is reported in \tabref{tab:hpo}, and its performance on the generalization set is displayed in \figref{fig:best_genset_performance}.


Comparisons to a hyper-parameter-optimized MLP, another optimized 1-D convolutional neural network (CNN), a likewise optimized LPTN, and a simple ordinary least squares (OLS) algorithm can be found in \tabref{tab:performance}.
All numbers are compiled from an evaluation on the generalization set.
Note that for pure ML algorithms the increased feature entries are to be computed in real-time as well rather than being measured, which, next to the model size, represents an additional computational demand on the embedded system.

\begin{table}[ht]
	\caption{State-of-the-art thermal modeling performance metrics as evaluated on the generalization set}
	\label{tab:performance}
	\centering
	\begin{tabular}{l|c|c|c|c| c}
	\hline\hline
	 	& \makecell{MSE\\(\si{\kelvin\squared})} & \makecell{$\ell_\infty$\\(\si{\kelvin})} & \makecell{LUT\\entries} & \makecell{Feature\\entries} & \makecell{Model\\size} \\ \hline \T
	 MLP \cite{KiWaBo2021_benchmark}	& 	5.58	 &	14.29	& 0		& 81		&  1444 \\
	 OLS \cite{KiWaBo2021_benchmark} &	4.47 &  9.85 	& 0 		& 81 	&  328 \\
	 CNN \cite{KiWaBo2020} 			&	4.43 &	15.54	& 0		& 81		&  4916 \\
	 LPTN \cite{WaBo2016}	  		& 	3.64 & 	7.37 	& 12810	& 4 		&  $\bm{34}$\\
	 TNN (small)						&   3.18 &	$\bm{5.84}$& 0		& 5 	&  64 \\
	 TNN (HPO optimum)				&	$\bm{2.87}$	 &	6.02 	& 0 & 5 	&  1525 \\
	 \hline\hline
	\end{tabular}
\end{table}	


\subsection{Model Size Pareto Front}
\begin{figure}[htb]
    \centering
    \includegraphics[width=0.49\textwidth]{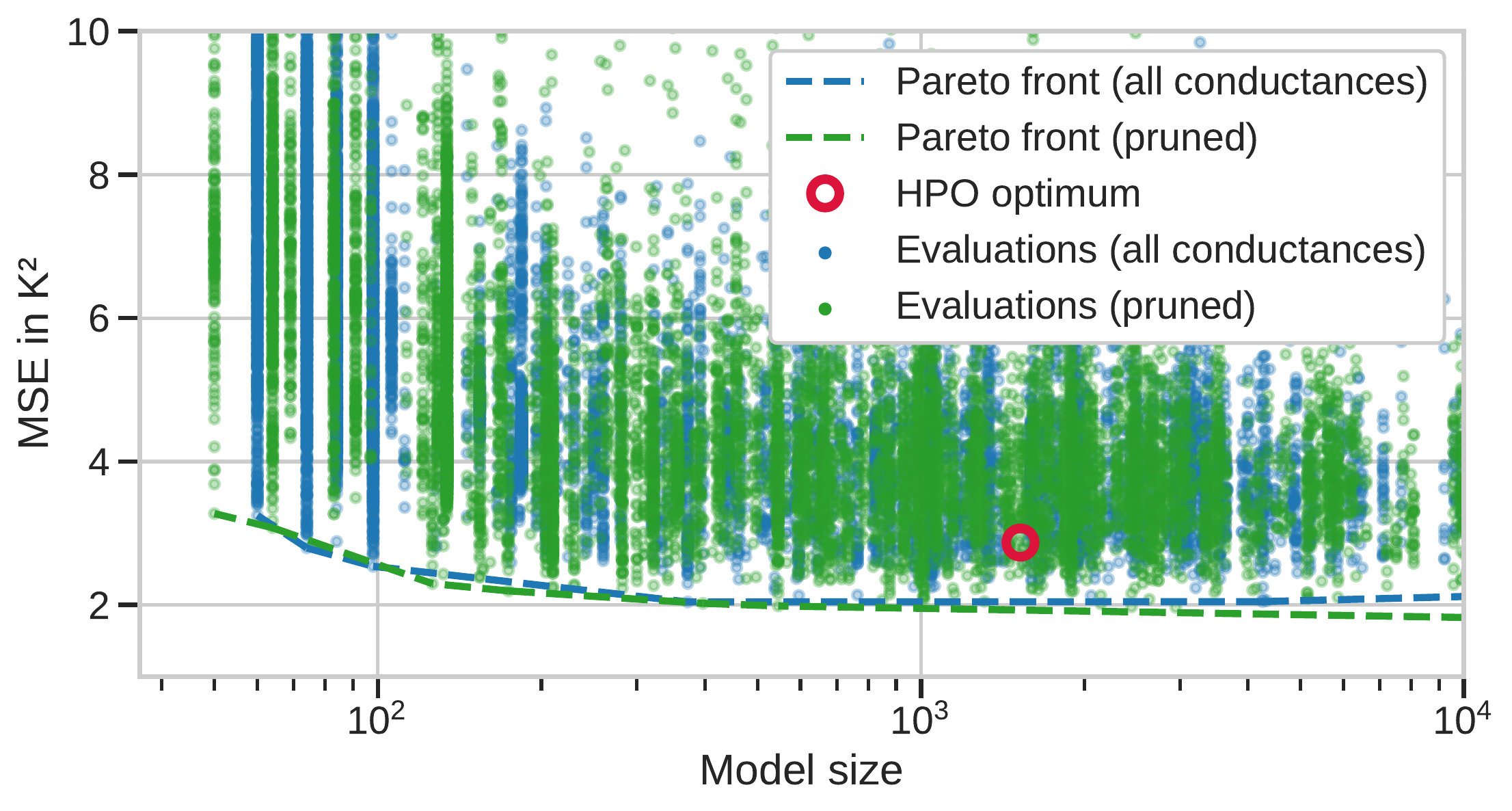}
    \caption{The MSE on the generalization set over different topological settings is illustrated. Here, $L_{\{\pi,\gamma\}}$ and $n_{\{\pi,\gamma\}}^{(l)}$ are independently sampled for both, $\bm\gamma$ and $\bm\pi$.}
    \label{fig:pareto}
\end{figure}

The foregoing HPO was guided by estimation accuracy alone, without considering model size.
However, in real-world applications, real-time capability and, thus, small, lightweight models play an important role as they are to be deployed prevalently on best-cost hardware.
In the following, a random search over a grid of a TNN's topological settings is conducted, keeping other hyperparameters at the optimum, in order to shape an informative Pareto front.
This shall give an overview of the trade-off between model size and estimation accuracy, see \figref{fig:pareto}.
The amount of layers and the number of units per layer for both, $\bm\gamma$ and $\bm\pi$, vary from $1$ to $3$ and $1$ to $128$, respectively, and independently from each other.
The number of units is varied exponentially in order to bound the combinatorial grid size growth.

It is evident that for very small model sizes the accuracy degrades only insignificantly.
Thus, the found hyperparameters reveal a robust guidance on the design of TNNs that are less influenced by the model size.
A topology with $2$ neurons in an intermediate layer for both thermal parameter approximators, which results in just below $100$ model parameters, seems to be a good trade-off between accuracy and model size.

It might appear incoherent that the kernel-based HPO optimum does not touch the Pareto front of the grid search.
This has two simple reasons: First, the foregoing HPO had to sample from a substantially larger sample space of $26$ hyperparameters, while the grid search sampled from just two, such that the over $2000$ candidates could have easily missed better local optima.
Second, the TPE sampling method was guided by the averaged estimation accuracy on two test sets, that might expose different peculiarities than the generalization set.


\subsection{Thermal Conductance Sparsification}
\label{sec:pruning}
\begin{figure}[htb]
    \centering
    \includegraphics[width=0.35\textwidth]{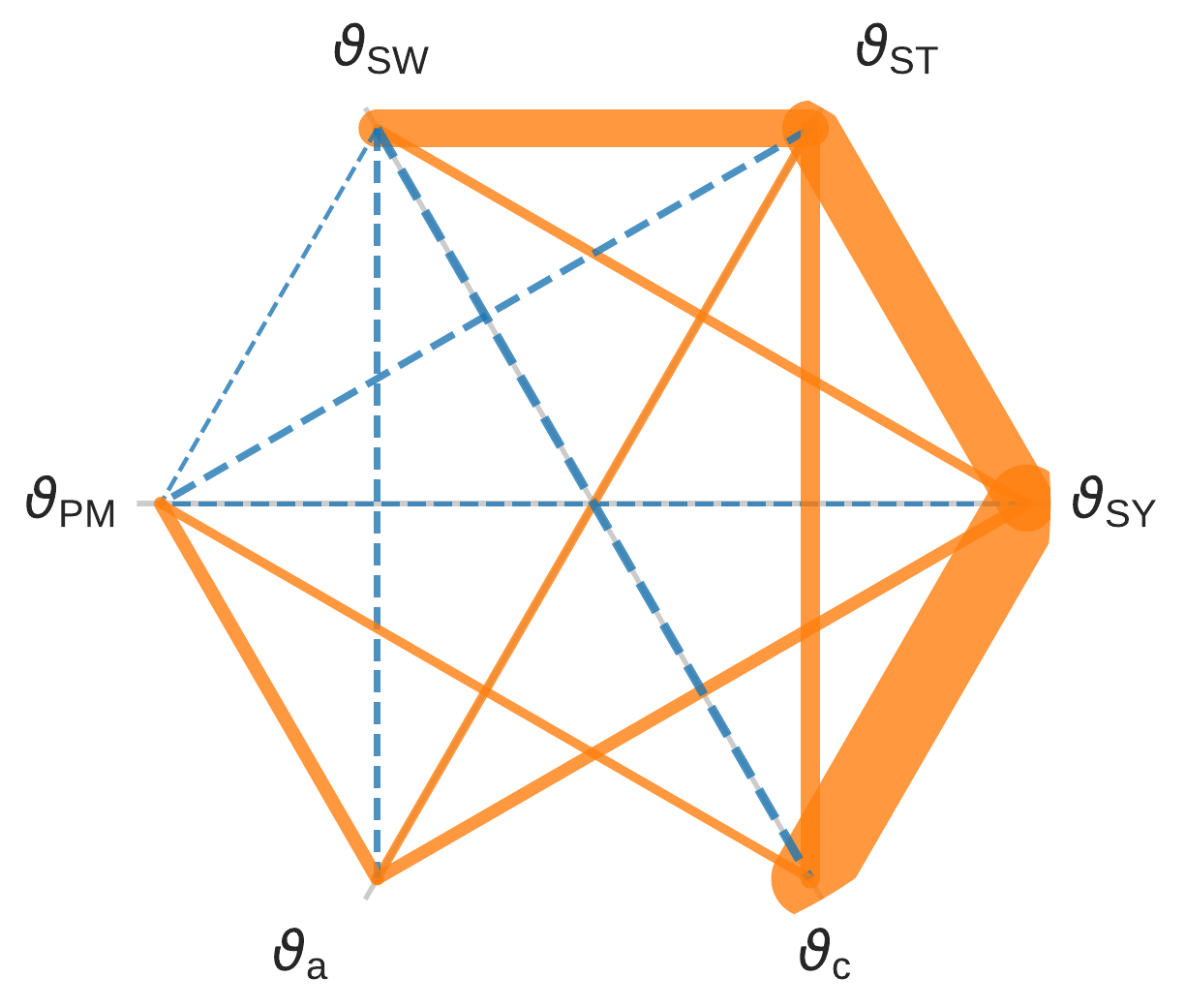}
    \caption{The median conductance values as being output on random input, averaged over $\bm\gamma$ networks derived from all successful experiments (i.e., $\text{MSE} < \SI{5}{\kelvin\squared}$) are visualized. Line thickness scales with conductance values linearly. Conductances under a certain threshold are drawn in blue, dashed lines, and are subject to pruning.}
    \label{fig:star}
\end{figure}

Having iterated over many TNN design parameters, a vast source for conductance analysis is available.
In an attempt to further reduce the model size, together with the fact that not all components within an intricate system necessarily offer direct heat paths to each other, the median activation of each conductance is investigated.
\figref{fig:star} shows an inter-conductance plot between all available temperatures, where line widths denote the median across a subset of all compiled $\bm\gamma$ networks' median output, when exposed to a random uniform input $\bm \phi \in \mathbb{R}^{n+o}$ with intervals $[0, 1.3]$.
Only those models that achieved an average MSE of below \SI{5}{\kelvin\squared} across the components are considered.
Those connections illustrated in blue, dashed lines are candidates for pruning.

It becomes apparent, that the highest thermal conductances exist between the coolant and stator temperatures, peaking between the stator yoke and coolant.
This is physically plausible as those two components are spatially attached and form a substantial, convective heat transfer (see \figref{fig:motor_schnitt}).
The weakest connections exist for the ambient and rotor temperature to all other, likewise.
The marginally significant ambient temperature conductance can be explained by its remote operating area.
However, the small conductances around the permanent magnet temperatures are counter-intuitive at first, but become more plausible under the fact that temperatures in the rotor are more power-loss-driven, and, thus, rather determined by internal heat generation than by conductive or convective heat transfer.
This hypothesis, if it were not verified in literature yet, is further corroborated by this temperature's small time constant -- best observed during cooling phases (refer to \figref{fig:best_genset_performance}).

The conductance analysis represents one of many benefits a TNN exhibits.
Without prior knowledge of the geometry or material in the system, the TNN learns the physical thermal relationship between measured nodes solely from data.

With the inferred conductance relevance, those under a certain threshold are pruned thereafter, and another random search over topological settings is conducted and shown in \figref{fig:pareto}.
The minimum amount of model parameters reduces from $60$ to $50$ due to conductance pruning, which is the case for one hidden layer of one neuron before each of the output layers of $\bm\gamma$ and $\bm\pi$.
Note that this can be further reduced through feature selection, which is out of this work's scope.
Keeping the model size constant, sparsification also has a beneficial effect on accuracy, because more parameters can be used to model fewer non-linear function behaviors. 


%

\subsection{Detuned Initial Conditions}
\label{sec:detuned_init}
Another advantage of TNNs is the possibility to set the initial condition through the cell state, which is not straight forward for usual ML methods.
Determining the initial condition is trivial in simulations and approximately equal to the ambient temperature in real-world scenarios.
However, if this initial guess is inaccurate, how long does it take to recover to a decent estimate?

\begin{figure}[htb]
    \centering
    \includegraphics[width=0.49\textwidth]{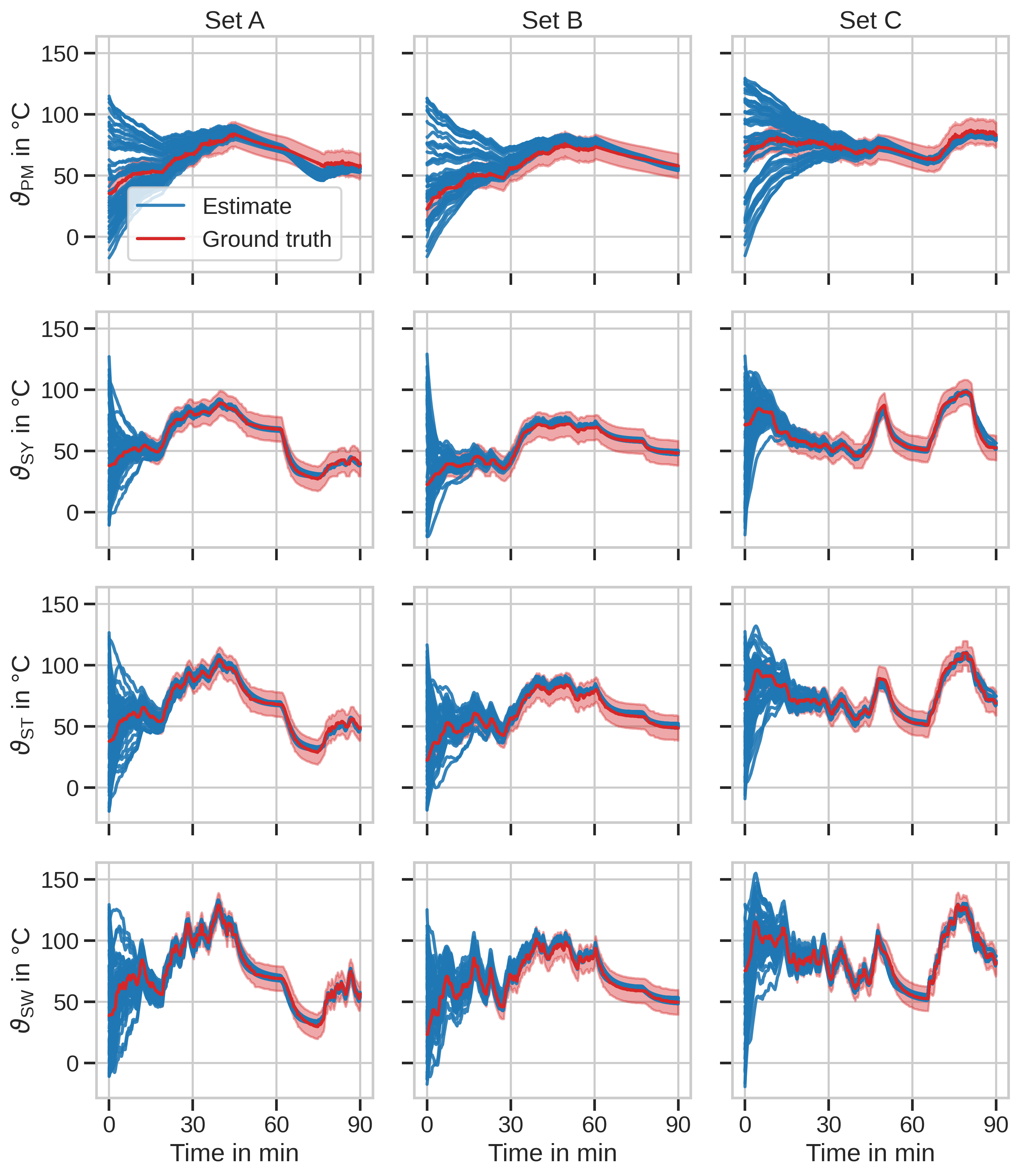}
    \caption{The impact of misconfigured initial conditions on estimation trajectories is demonstrated. The ground truth trajectory is accompanied by a red-shaded \SI{\pm 10}{\degreeCelsius} error band.}
    \label{fig:detuned_init}
\end{figure}

\figref{fig:detuned_init} exemplifies different scenarios with varying detuned initial estimates.
It becomes evident that the worst-case estimate recovers to a \SI{\pm 10}{\degreeCelsius} error band (red-shaded area) in under \SI{30}{\minute} for the stator temperatures, and in under \SI{45}{\minute} for the rotor temperature.
The higher the thermal capacitance of a component, the slower the convergence is to be expected.
This is acceptable for most applications considering the low risk of a flawed initial condition approximation.
Similar observations can be made for pure LPTN models.

Finally, it can be noted that a TNN, being quasi-linear in its state-space representation, lends itself as internal model for state observers (Kalman, Luenberger) as soon as parts of the target temperatures are (virtually) measured in a field application \cite{WaBo17_fusion, Gaona2020}.
Standard observer types for linear systems would already suffice while the internal system model is updated by the TNN.
Given the external feedback, deviations during estimation and at initialization could be reduced substantially more rapidly.

\section{Conclusion}
A hybrid thermal model for real-time temperature estimation, the so-called thermal neural network, is presented in this work, which exhibits the benefits of both worlds -- consolidated knowledge on heat transfer phenomena and nonlinear, data-driven function approximation with ML tools.
Due to a TNN's outer LPTN structure, which lifts the black-box model to a quasi-LPV model, it is applicable everywhere an LPTN would be seen appropriate, but with the additional adaptivity to parasitic real-world effects as observed in measurement data.
For the same reason, initial conditions can be set, and model states are physically interpretable, giving rise to conclusions over the thermal characteristics of the measured system.
Being a composite of three general-purpose function approximators, a TNN is end-to-end differentiable such that modern automatic differentiation frameworks are employable.
Neither expert knowledge nor motor sheet data is necessary to design a TNN, but they can help streamlining its structure.
Having these properties combined in one model is unique throughout the thermal modeling literature to the authors' best knowledge.

One disadvantage a TNN shares with common ML approaches is the dependence on measurement data, although transferring knowledge from one trained TNN model to another likewise manufactured system is indeed possible.
Moreover, data is more and more available even in power system domains nowadays.

Future investigations shall consider applying the UDE and SSNN principle to other domains of energy conversion where latent states are estimated in real-time, e.g., battery systems' state of charge, or induction motors' differential inductances and magnetic flux linkage.
Applications to reinforcement learning controllers - especially in power systems - are obvious and promising as well.

\bibliographystyle{IEEEtran}
\bibliography{../../references/library}

\begin{thebibliography}{10}
\providecommand{\url}[1]{#1}
\csname url@samestyle\endcsname
\providecommand{\newblock}{\relax}
\providecommand{\bibinfo}[2]{#2}
\providecommand{\BIBentrySTDinterwordspacing}{\spaceskip=0pt\relax}
\providecommand{\BIBentryALTinterwordstretchfactor}{4}
\providecommand{\BIBentryALTinterwordspacing}{\spaceskip=\fontdimen2\font plus
\BIBentryALTinterwordstretchfactor\fontdimen3\font minus
  \fontdimen4\font\relax}
\providecommand{\BIBforeignlanguage}[2]{{%
\expandafter\ifx\csname l@#1\endcsname\relax
\typeout{** WARNING: IEEEtran.bst: No hyphenation pattern has been}%
\typeout{** loaded for the language `#1'. Using the pattern for}%
\typeout{** the default language instead.}%
\else
\language=\csname l@#1\endcsname
\fi
#2}}
\providecommand{\BIBdecl}{\relax}
\BIBdecl

\bibitem{IaLu2014}
M.~Iachello, V.~{De Luca}, G.~Petrone, N.~Testa, L.~Fortuna, G.~Cammarata,
  S.~Graziani, and M.~Frasca, ``{Lumped Parameter Modeling for Thermal
  Characterization of High-Power Modules},'' \emph{IEEE Transactions on
  Components, Packaging and Manufacturing Technology}, vol.~4, no.~10, pp.
  1613--1623, 2014.

\bibitem{BaMaGhi2016}
A.~S. Bahman, K.~Ma, P.~Ghimire, F.~Iannuzzo, and F.~Blaabjerg, ``{A 3-D-Lumped
  Thermal Network Model for Long-Term Load Profiles Analysis in High-Power IGBT
  Modules},'' \emph{IEEE Journal of Emerging and Selected Topics in Power
  Electronics}, vol.~4, no.~3, pp. 1050--1063, 2016.

\bibitem{WaBo2016}
O.~Wallscheid and J.~B{\"{o}}cker, ``{Global Identification of a Low-Order
  Lumped-Parameter Thermal Network for Permanent Magnet Synchronous Motors},''
  \emph{IEEE Transactions on Energy Conversion}, vol.~31, no.~1, pp. 354--365,
  2016.

\bibitem{BraHe2012}
N.~Bracikowski, M.~Hecquet, P.~Brochet, and S.~V. Shirinskii, ``{Multiphysics
  Modeling of a Permanent Magnet Synchronous Machine by Using Lumped Models},''
  \emph{IEEE Transactions on Industrial Electronics}, vol.~59, no.~6, pp.
  2426--2437, 2012.

\bibitem{BoCaCo2015}
A.~Boglietti, E.~Carpaneto, M.~Cossale, and S.~Vaschetto, ``{Stator-Winding
  Thermal Models for Short-Time Thermal Transients: Definition and
  Validation},'' \emph{IEEE Transactions on Industrial Electronics}, vol.~63,
  no.~5, pp. 2713--2721, 2016.

\bibitem{RaGoEa2013}
A.~P. Ramallo-Gonz{\'{a}}lez, M.~E. Eames, and D.~A. Coley, ``{Lumped parameter
  models for building thermal modelling: An analytic approach to simplifying
  complex multi-layered constructions},'' \emph{Energy and Buildings}, vol.~60,
  pp. 174--184, 2013.

\bibitem{BoCa2009}
A.~Boglietti, A.~Cavagnino, D.~Staton, M.~Shanel, M.~Mueller, and C.~Mejuto,
  ``{Evolution and Modern Approaches for Thermal Analysis of Electrical
  Machines},'' \emph{IEEE Transactions on Industrial Electronics}, vol.~56,
  no.~3, pp. 871--882, 2009.

\bibitem{ReiFeYo2015}
D.~D. Reigosa, D.~Fernandez, H.~Yoshida, T.~Kato, and F.~Briz,
  ``{Permanent-Magnet Temperature Estimation in PMSMs Using Pulsating
  High-Frequency Current Injection},'' \emph{IEEE Transactions on Industry
  Applications}, vol.~51, no.~4, pp. 3159--3168, 2015.

\bibitem{ReiFeMa2019}
D.~Reigosa, D.~Fern{\'{a}}ndez, M.~Mart{\'{i}}nez, J.~M. Guerrero, A.~B. Diez,
  and F.~Briz, ``{Magnet Temperature Estimation in Permanent Magnet Synchronous
  Machines Using the High Frequency Inductance},'' \emph{IEEE Transactions on
  Industry Applications}, vol.~55, no.~3, pp. 2750--2757, 2019.

\bibitem{SpeWaBo2014}
O.~Wallscheid, A.~Specht, and J.~B{\"{o}}cker, ``{Observing the
  Permanent-Magnet Temperature of Synchronous Motors Based on Electrical
  Fundamental Wave Model Quantities},'' \emph{IEEE Transactions on Industrial
  Electronics}, vol.~64, no.~5, pp. 3921--3929, 2017.

\bibitem{WaHuPe16}
O.~Wallscheid, T.~Huber, W.~Peters, and J.~B{\"{o}}cker, ``{A Critical Review
  of Techniques to Determine the Magnet Temperature of Permanent Magnet
  Synchronous Motors Under Real-Time Conditions},'' \emph{EPE Journal},
  vol.~26, pp. 1--10, 2016.

\bibitem{GeWaBo2020}
E.~Gedlu, O.~Wallscheid, and J.~B{\"{o}}cker, ``{Permanent Magnet Synchronous
  Machine Temperature Estimation using Low-Order Lumped-Parameter Thermal
  Network with Extended Iron Loss Model},'' TechRxiv e-prints: 11671401, 2020.

\bibitem{QiScheDo2014}
F.~Qi, M.~Schenk, and R.~W. {De Doncker}, ``{Discussing details of lumped
  parameter thermal modeling in electrical machines},'' in \emph{7th IET
  International Conference on Power Electronics, Machines and Drives (PEMD
  2014)}, 2014, pp. 1--6.

\bibitem{WoeBraDre2020}
D.~W{\"{o}}ckinger, G.~Bramerdorfer, S.~Drexler, S.~Vaschetto, A.~Cavagnino,
  A.~Tenconi, W.~Amrhein, and F.~Jeske, ``{Measurement-Based Optimization of
  Thermal Networks for Temperature Monitoring of Outer Rotor PM Machines},'' in
  \emph{2020 IEEE Energy Conversion Congress and Exposition (ECCE)}, 2020, pp.
  4261--4268.

\bibitem{KiWaBo2020}
W.~Kirchg{\"{a}}ssner, O.~Wallscheid, and J.~B{\"{o}}cker, ``{Estimating
  Electric Motor Temperatures with Deep Residual Machine Learning},''
  \emph{IEEE Transactions on Power Electronics}, vol.~36, no.~7, pp.
  7480--7488, 2020.

\bibitem{KiWaBo2021_benchmark}
------, ``{Data-Driven Permanent Magnet Temperature Estimation in Synchronous
  Motors with Supervised Machine Learning: A Benchmark},'' \emph{IEEE
  Transactions on Energy Conversion}, 2021.

\bibitem{LeeHa2020}
J.~Lee and J.~Ha, ``{Temperature Estimation of PMSM Using a
  Difference-Estimating Feedforward Neural Network},'' \emph{IEEE Access},
  vol.~8, pp. 130\,855--130\,865, 2020.

\bibitem{ZhaGuOg2018}
K.~Zhang, A.~Guliani, S.~Ogrenci-Memik, G.~Memik, K.~Yoshii, R.~Sankaran, and
  P.~Beckman, ``{Machine Learning-Based Temperature Prediction for Runtime
  Thermal Management Across System Components},'' \emph{IEEE Transactions on
  Parallel and Distributed Systems}, vol.~29, no.~2, pp. 405--419, 2018.

\bibitem{Bergman2007}
T.~L. Bergman, F.~P. Incropera, D.~P. DeWitt, and A.~S. Lavine,
  \emph{{Fundamentals of Heat and Mass Transfer}}, 6th~ed.\hskip 1em plus 0.5em
  minus 0.4em\relax John Wiley {\&} Sons, 2007.

\bibitem{BoCaPa06}
A.~Boglietti, A.~Cavagnino, M.~Parvis, and A.~Vallan, ``{Evaluation of
  Radiation Thermal Resistances in Industrial Motors},'' \emph{IEEE
  Transactions on Industry Applications}, vol. Vol. 42, N, pp. 688--693, 2006.

\bibitem{HiChiHo12}
D.~A. Howey, P.~R.~N. Childs, and A.~S. Holmes, ``{Air-Gap Convection in
  Rotating Electrical Machines},'' \emph{IEEE Transactions on Industrial
  Electronics}, vol. Vol. 59, N, pp. 1367--1375, 2012.

\bibitem{GunesBaydin2018}
A.~G. Baydin, B.~A. Pearlmutter, A.~Radul, and J.~M. Siskind, ``{Automatic
  differentiation in machine learning: A survey},'' \emph{Journal of Machine
  Learning Research}, vol.~18, pp. 1--43, 2018.

\bibitem{Rudy2017}
S.~H. Rudy, S.~L. Brunton, J.~L. Proctor, and J.~N. Kutz, ``{Data-Driven
  Discovery of Partial Differential Equations},'' \emph{Science Advances},
  vol.~3, no.~4, p. e1602614, 2017.

\bibitem{CheRuBeDu2018}
R.~T.~Q. Chen, Y.~Rubanova, J.~Bettencourt, and D.~K. Duvenaud, ``{Neural
  Ordinary Differential Equations},'' in \emph{Advances in Neural Information
  Processing Systems}, vol.~31.\hskip 1em plus 0.5em minus 0.4em\relax Curran
  Associates, Inc., 2018, pp. 6571--6583.

\bibitem{RaPeKa2019}
M.~Raissi, P.~Perdikaris, and G.~E. Karniadakis, ``{Physics-Informed Neural
  Networks: A Deep Learning Framework for Solving Forward and Inverse Problems
  Involving Nonlinear Partial Differential Equations},'' \emph{Journal of
  Computational Physics}, vol. 378, pp. 686--707, 2019.

\bibitem{Rackauckas2020}
C.~Rackauckas, Y.~Ma, J.~Martensen, C.~Warner, K.~Zubov, R.~Supekar,
  D.~Skinner, and A.~Ramadhan, ``{Universal Differential Equations for
  Scientific Machine Learning},'' ArXiv e-prints: 2001.04385[cs.LG], 2020.

\bibitem{HoSti1989}
K.~Hornik, M.~Stinchcombe, and H.~White, ``{Multilayer Feedforward Networks are
  Universal Approximators},'' \emph{Neural Networks}, vol.~2, no.~5, pp.
  359--366, 1989.

\bibitem{Rivals1996}
I.~Rivals and L.~Personnaz, ``{Black-Box Modeling With State-Space Neural
  Networks},'' in \emph{Neural Adaptive Control Technology}.\hskip 1em plus
  0.5em minus 0.4em\relax World Scientific, 1996, pp. 237--264.

\bibitem{ssNN98}
J.~M. Zamarre{\~{n}}o and P.~Vega, ``{State Space Neural Network. Properties
  and Application},'' \emph{Neural Networks}, vol.~11, no.~6, pp. 1099--1112,
  1998.

\bibitem{AbbWe2008}
H.~Abbas and H.~Werner, ``{Polytopic Quasi-LPV Models Based on Neural
  State-Space Models and Application to Air Charge Control of a SI Engine},''
  \emph{IFAC Proceedings Volumes}, vol.~41, no.~2, pp. 6466--6471, 2008.

\bibitem{PaMi2012}
R.~Pascanu, T.~Mikolov, and Y.~Bengio, ``{Understanding the Exploding Gradient
  Problem},'' \emph{Proceedings of The 30th International Conference on Machine
  Learning}, 2012.

\bibitem{GeSch1999}
F.~A. Gers and F.~Cummins, ``{Learning to Forget: Continual Prediction with
  LSTM},'' in \emph{Ninth International Conference on Artificial Neural
  Networks}, vol.~2, 1999, pp. 1--19.

\bibitem{WiPe90}
R.~J. Williams and J.~Peng, ``{An Efficient Gradient-Based Algorithm for
  On-line Training of Recurrent Network Trajectories},'' \emph{Neural
  computation}, vol.~2, no.~4, pp. 490--501, 1990.

\bibitem{KiWaBo2019_2}
W.~Kirchg{\"{a}}ssner, O.~Wallscheid, and J.~B{\"{o}}cker, ``{Empirical
  Evaluation of Exponentially Weighted Moving Averages for Simple Linear
  Thermal Modeling of Permanent Magnet Synchronous Machines},'' in
  \emph{Proceedings of the 28th International Symposium on Industrial
  Electronics}, 2019, pp. 318--323.

\bibitem{tf2}
M.~Abadi, P.~Barham, J.~Chen, Z.~Chen, A.~Davis, J.~Dean, M.~Devin,
  S.~Ghemawat, G.~Irving, M.~Isard, and Others, ``{Tensorflow: A System for
  Large-Scale Machine Learning},'' in \emph{12th USENIX symposium on operating
  systems design and implementation (OSDI 16)}, 2016, pp. 265--283.

\bibitem{optuna_2019}
T.~Akiba, S.~Sano, T.~Yanase, T.~Ohta, and M.~Koyama, ``{Optuna: A
  Next-generation Hyperparameter Optimization Framework},'' in
  \emph{Proceedings of the 25rd ACM SIGKDD International Conference on
  Knowledge Discovery and Data Mining}, 2019.

\bibitem{BeBaBe2011}
J.~Bergstra, R.~Bardenet, Y.~Bengio, and B.~K{\'{e}}gl, ``{Algorithms for
  Hyper-Parameter Optimization},'' in \emph{25th annual conference on neural
  information processing systems (NIPS 2011)}, vol.~24.\hskip 1em plus 0.5em
  minus 0.4em\relax Neural Information Processing Systems Foundation, 2011.

\bibitem{WaBo17_fusion}
O.~Wallscheid and J.~B{\"{o}}cker, ``{Fusion of Direct and Indirect Temperature
  Estimation Techniques for Permanent Magnet Synchronous Motors},'' in
  \emph{2017 IEEE International Electric Machines and Drives Conference
  (IEMDC)}, 2017, pp. 1--8.

\bibitem{Gaona2020}
D.~Gaona, O.~Wallscheid, and J.~B{\"{o}}cker, ``{Improved Fusion of Permanent
  Magnet Temperature Estimation Techniques for Synchronous Motors Using a
  Kalman Filter},'' \emph{IEEE Transactions on Industrial Electronics},
  vol.~67, no.~3, pp. 1708--1717, 2020.

\end{thebibliography}

\end{document}